\pdfoutput=1

\documentclass[11pt]{article}

\usepackage[]{acl}

\usepackage{times}
\usepackage{latexsym}
\usepackage{graphicx}
\usepackage{svg}
\usepackage{amsmath}
\usepackage{verbatim}
\usepackage{booktabs}
\usepackage{multirow}
\usepackage{amsfonts}
\usepackage{subcaption}
\usepackage[ruled, noend]{algorithm2e}
\usepackage{float}

\usepackage[T1]{fontenc}

\usepackage[utf8]{inputenc}

\usepackage{microtype}

\usepackage{fdsymbol}

\title{Crake: Causal-Enhanced Table-Filler for Question Answering\\ over Large Scale Knowledge Base}

\author{Minhao Zhang$^{1\clubsuit}$ \quad Ruoyu Zhang$^{1\clubsuit}$ \quad Yanzeng Li$^{1\vardiamondsuit}$ \quad Lei Zou$^{1,2\clubsuit}$\\
$^1$
Wangxuan Institute of Computer Technology (WICT), 
Peking University, China; \\ $^2$Beijing Academy of Artificial Intelligence, Beijing, China;\\
\texttt{$^{\clubsuit}$\{zhangminhao,ry\_zhang,zoulei\}@pku.edu.cn} \\
\texttt{$^{\vardiamondsuit}$ liyanzeng@stu.pku.edu.cn} \\
}

\begin{document}
\maketitle
\begin{abstract}
Semantic parsing solves knowledge base (KB) question answering (KBQA) by composing a KB query, which generally involves node extraction (NE) and graph composition (GC) to detect and connect related nodes in a query.
Despite the strong causal effects between NE and GC, previous works fail to directly model such causalities in their pipeline, hindering the learning of subtask correlations. Also, the sequence-generation process for GC in previous works induces ambiguity and exposure bias, which further harms accuracy. In this work, we formalize semantic parsing into two stages. In the first stage (graph structure generation), we propose a causal-enhanced table-filler to overcome the issues in sequence-modelling and to learn the internal causalities. In the second stage (relation extraction), an efficient beam-search algorithm is presented to scale complex queries on large-scale KBs. Experiments on LC-QuAD 1.0 indicate that our method surpasses previous state-of-the-arts by a large margin (17\%) while remaining time and space efficiency. The code and models are available at \url{https://github.com/AOZMH/Crake}.

\end{abstract}

\section{Introduction}
\label{sec:intro}
To incorporate knowledge in real-world question-answering systems, knowledge base question answering (KBQA) utilizes a background knowledge base (KB) as the source of answers to factoid natural language questions. Leveraging the versatility of KB query languages like SPARQL \citep{prud2008sparql}, many previous works \citep{unger2012template, yahya2012natural} adopted a semantic parsing paradigm for KBQA, in which questions are converted to equivalent SPARQL queries and answers are given by executing the queries in KB. Regarding the intrinsic graph structure of SPARQLs, some works further reduced such procedure as generating the query graph of SPARQLs w.r.t. questions. However, these methods either require auxiliary tools (e.g. AMR in \citealp{kapanipathi2021leveraging}, constituency tree in \citealp{hu2021edg}, dependency tree in \citealp{hu2017answering}) causing potential cascading errors, or rely on predefined templates \citep{cui2019kbqa, athreya2021template} limiting their expressiveness and generalization abilities.

\begin{figure}[t]
    \centering
    \includegraphics[width=1.0\linewidth]{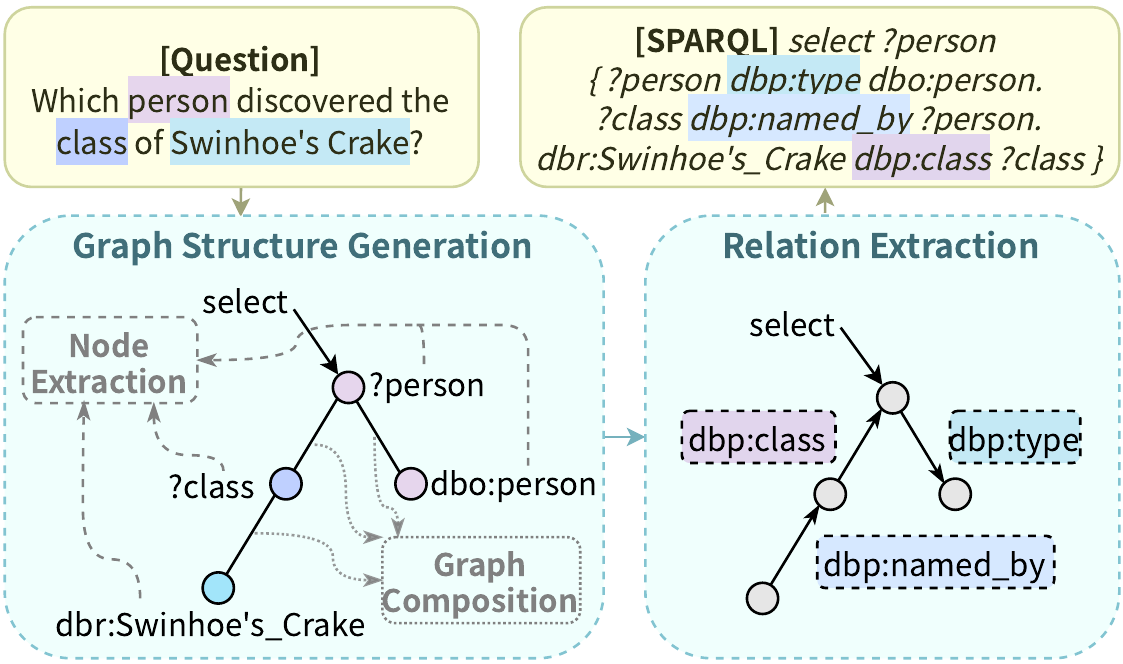}
    \caption{Generating a query graph (bottom) by two stages to represent the SPARQL (right-top). At graph structure generation stage, node-extraction generates all graph nodes while graph-composition adds unlabeled edges between proper nodes. Then, the relation extraction stage decides the specific predicate of each edge.}
    \label{fig:two_stage}
\end{figure}

To address these, efforts were made on devising independent pipelines for query graph construction \citep{lin2021deep}. As in Figure \ref{fig:two_stage}, these pipelines usually involve a node extraction (NE) module to detect the mentions of all nodes in query graph and link entity mentions, a graph composition (GC) module to connect related nodes given by NE, and a relation extraction (RE) module deciding the KB predicate corresponding to each edge added in GC. In this framework, two drawbacks exist in previous works: 1) we observe strong causal effects between NE and GC, e.g. edges connected by GC are valid only between the node mentions extracted in NE, making GC decisions highly dependent on NE. To this regard, previous works \citep{zhang2021namer, ravishankar2021two} that perform NE and GC separately without causal-modelling may fall short in deeply comprehending the correlated tasks and accurately generating query graphs. 2) GC is commonly modelled as a sequence-generation in prior methods, either through generative decoder \citep{shen2019multi, chen2021outlining} or via stage-transition \citep{yih2015semantic, hu2018state}. However, sequence-modelling generally undergoes sequence ambiguity and exposure bias \citep{zhang-etal-2019-bridging} that harms model accuracy.

In this work, we formalize the generation of query graph in a two-staged manner as in Figure \ref{fig:two_stage}. At the first stage, we tackle the aforesaid weaknesses by a novel causal-enhanced table-filling model to jointly complete NE and GC, resulting in a query graph structure representing the connectivity of all nodes. More specifically, inspired by \citet{chen-etal-2020-exploring-logically}, we utilize a label transfer mechanism to facilitate the acquisition of causality between NE and GC (which solves drawback 1 above). Further, we apply a table-filler to decode all edges simultaneously, which naturally circumvents the ambiguity and bias of iterative decoding (and solves drawback 2). For the second stage, we propose a beam-search-based relation extraction algorithm to determine the predicate that binds to each graph edge. Differ from prior works, we perform candidate predicate retrieval and ranking alternately for each edge, limiting the candidate scale linearly w.r.t. KB degree and making the algorithm scalable for large-scale KBs like DBpedia.

In short, the major contributions of this paper are: 1) to our knowledge, we are the first to model GC as a table-filling process, which prevents the ambiguity and bias in prior works; 2) we model the intrinsic causal effects in KBQA to grasp subtask correlations and improve pipeline integrity; 3) our method outperforms previous state-of-the-arts on LC-QuAD 1.0, a prominent KBQA benchmark, by a large margin ($\sim\!\!17\%$), further experiments verifies the effectiveness of our approach.

\section{Preliminaries}
\subsection{Problem Setting}
We solve KBQA in a semantic parsing way, given a question (left-top in Figure \ref{fig:two_stage}), we generate a SPARQL query (right-top in Figure \ref{fig:two_stage}) to represent its semantics and answer the question by executing the query in KB. By definition, SPARQL describes a query graph with each triple in its body referring to a graph edge; by matching the graph pattern in KB, certain KB entries binding to the query graph can be processed as query results (e.g. in Table \ref{tab:trigger_words} for SELECT queries, all entries binding to the "select" node are results; for JUDGE queries, the existence of matched entries determines the boolean result). Hence, our task is further specified as constructing the query graph (bottom of Figure \ref{fig:two_stage}) of a question to represent its corresponding SPARQL.

\begin{table}[t]
\centering

\resizebox{0.8\columnwidth}{!}{
\begin{tabular}{c c}
    \toprule
    \bfseries Type & \bfseries Example SPARQL\\
    \cmidrule(lr){1-2}
    JUDGE & ask \{dbr:New\_York a dbo:City\} \\
    COUNT & select count(?x) \{?x a dbo:City\}  \\
    SELECT & select ?x \{?x a dbo:City\} \\
    \bottomrule
\end{tabular}}
\caption{Supported query types.}

\label{tab:trigger_words}
\end{table}

\subsection{Methodology Overview}
Illustrated by Figure \ref{fig:two_stage}, we construct the query graph in two stages. In the graph structure generation stage (bottom-left in Figure \ref{fig:two_stage}), we extract all graph nodes by finding the mention of each node in question and its tag among $\{ variable,entity,type \}$, e.g. the mention and tag for the node $?class$ is "class" and variable, respectively. Further, we link all non-variable nodes to KB entries, e.g. the $type$ node with mention "person" links to \textit{dbo:person} in Figure \ref{fig:two_stage}. Also, we decide the target ("select") node of the graph and add undirected edges between the nodes that are connected in the query graph, resulting in a graph structure representing the connectivity of all nodes.

Since all edges above are undirected and unlabeled, we fill in the exact KB predicate of each edge in the relation extraction stage (bottom-right in Figure \ref{fig:two_stage}) to construct a complete query graph. 

Finally, we compose a SPARQL w.r.t. the query graph as output. Note that the body of the SPARQL exactly corresponds to the query graph, so only the SPARQL header is yet undetermined. Like \citealp{hu2021edg}, we collect frequent trigger words in the train data to classify questions into COUNT, JUDGE or SELECT queries as in Table \ref{tab:trigger_words} (e.g. a question beginning with "is" triggers JUDGE). Thus, an entire SPARQL can now be formed. In the following sections, we expatiate our methodology for the two aforementioned stages.

\begin{figure*}[t]
    \centering
    \includegraphics[width=0.999\linewidth]{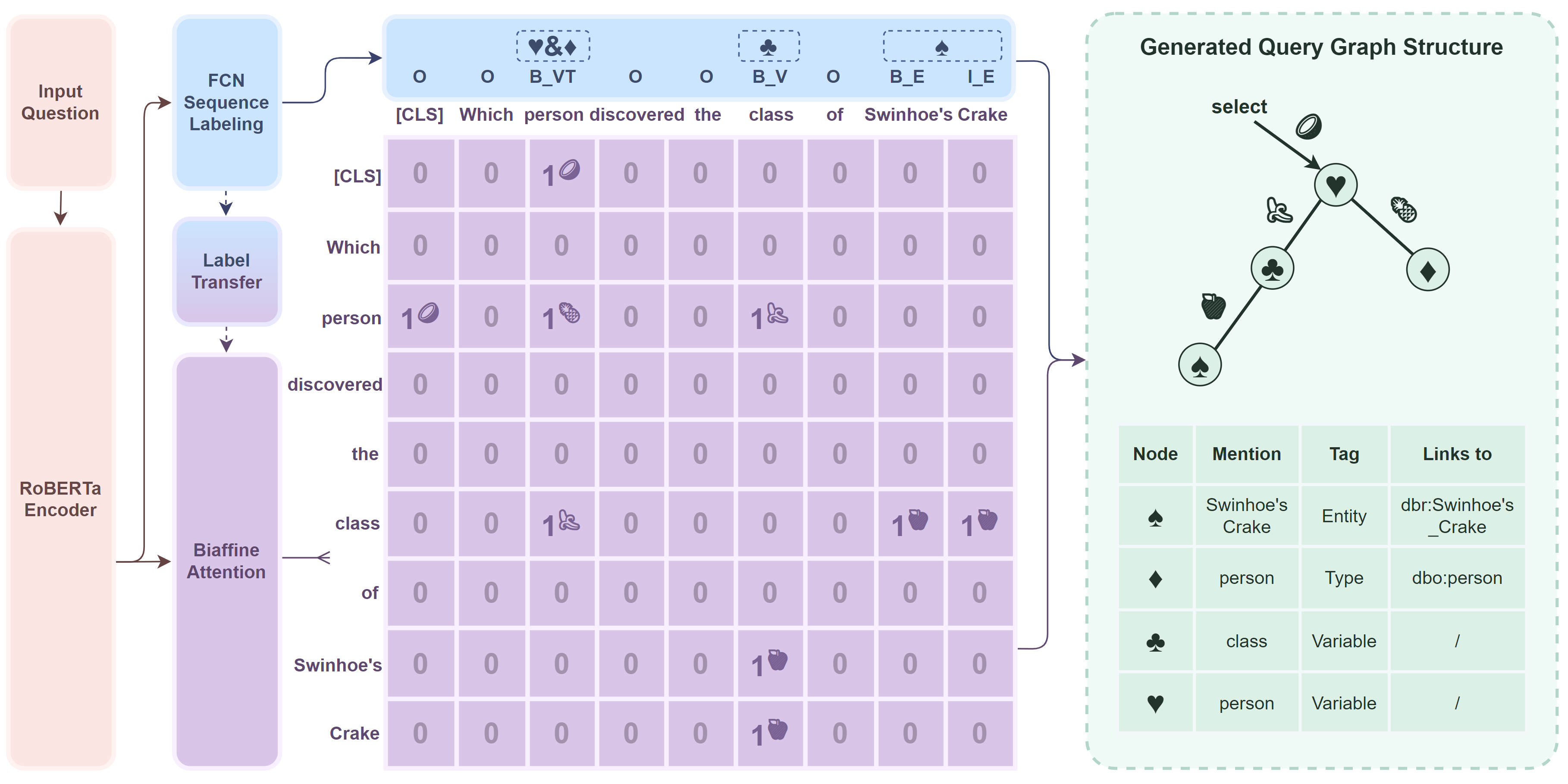}
    \caption{Causal-enhanced table-filling model for graph structure generation. The label-to-node and table-to-edge correspondence is illustrated by the poker and fruit symbols respectively.}
    \label{fig:graph_structure_generation}
\end{figure*}

\section{Graph Structure Generation (GSG)}
\label{sec:gsg}
The overview of the model proposed for graph structure generation is illustrated by Figure \ref{fig:graph_structure_generation}. As discussed in Section \ref{sec:intro}, the model jointly deals with node extraction and graph composition via causal-modelling, which is detailed in this section below.

\subsection{Node Extraction (NE)}
Node extraction discovers all nodes in the query graph, i.e. \{\textit{?person}, \textit{?class}, \textit{dbr:Swinhoe's\_Crake}, \textit{dbo:person}\} in Figure \ref{fig:two_stage}. We represent a node by its mention and tag, i.e. ("person", \textit{variable}), ("class", \textit{variable}), ("Swinhoe's Crake", \textit{entity}) and ("person", \textit{type}) for each node respectively.

This goal can naturally be achieved by multi-class sequence labeling. More specifically, let $\mathbf{Q}\in \mathbb{N}^{n}$ be the question (token ids) with length $n$, we first encode it into hidden features $\mathbf{H}_{rb}$ by a RoBERTa \citep{liu2019roberta} encoder $E_{rb}: \mathbb{N}^{n} \rightarrow \mathbb{R}^{n\times h_{rb}}$ with hidden size $h_{rb}$:
\[ \mathbf{H}_{rb} = E_{rb}(\mathbf{Q}) \in \mathbb{R}^{n\times h_{rb}} \]
Then, $\mathbf{H}_{rb}$ is projected by a fully-connected-network (FCN) $E_{ne}: \mathbb{R}^{n\times h_{rb}} \rightarrow \mathbb{R}^{n\times |L|}$ into $\mathbf{Y}_{ne}$ in label space:
\[ 
\mathbf{Y}_{ne} = E_{ne}(\mathbf{H}_{rb}) \in \mathbb{R}^{n\times |L|}
\]
$L=\{O\}\cup\{B,I\}\times\{V,E,T,VT\}$ is the label set denoting the mention span of variables (V), entities (E), types (T), or overlapping variable and type (VT). Now, the label prediction of each token can be given by $\mathbf{P}_{ne}=argmax(\mathbf{Y}_{ne})$; also, given the gold token labels $\mathbf{G}_{ne}\in \mathbb{N}^{n}$ (Figure \ref{fig:graph_structure_generation} top), a model for NE can be trained by optimizing:
\[ \ell_{ne} = -\frac{1}{n} \sum\limits_{i=1}^n log(\text{softmax}(\mathbf{Y}_{ne})[i; \mathbf{G}_{ne}[i]]) \]
Where $[\cdot ]$ denotes tensor indexing.

After detecting all node mentions and tags, we link each non-variable node to KB entries by DBpedia Lookup and a mention-to-type dictionary built on train data to align the graph structure with KB. See Appendix \ref{sec:appendix_linking} for more details in node linking.

\subsection{Graph Composition (GC)}
\label{sec:gc}
After node extraction, all nodes in the query graph remain unconnected. To form the structure of the query graph, graph composition inserts unlabeled and undirected edges between the nodes that are related in the query graph, leaving the specific predicate of each edge yet unresolved. Formerly, graph composition is commonly modelled as a edge-sequence-generation process via stage-transition \citep{yih2015semantic, hu2018state} or generative decoders \citep{shen2019multi, chen2021outlining}. Despite the strong expressiveness, modelling graph composition by a sequence usually suffers from two issues: 1) while the edge sequence is ordered, edges in the query graph are a set without order. For a graph with two edges $e_1$ and $e_2$, both sequence $e_1{\text -}e_2$ and $e_2{\text -}e_1$ correctly represents the edges in the graph, but they are distinct from the perspective of sequence-generation. As a result, the edge set itself becomes ambiguous for the sequence, which confuses the model when comprehending a sequence and potentially decelerates the convergence. 2) As discussed by \citealp{zhang-etal-2019-bridging}, without extra augmentation, sequence-generation generally endures an exposure bias between training and inference, harming the model's accuracy when predicting. Hence, a robust model should address the issues above properly.

Here, we model graph composition by a table-filling process to decide all edges simultaneously involving no sequence-generation, which naturally circumvents all issues above. Let $\mathbf{H}_{gc}\in \mathbb{R}^{n\times h_{gc}}$ be the hidden features for graph composition (the full definition of $\mathbf{H}_{gc}$ with causal-modelling is given in Section \ref{sec:causal_modelling}; without causal-modelling, we simply have $\mathbf{H}_{gc}=\mathbf{H}_{rb}$), we adopt a biaffine attention model \citep{dozat2016deep, wang2021unire} to convert $\mathbf{H}_{gc}$ into a table denoting the relationship between each token pair. More specifically, through two multi-layer-perceptrons (MLP) $E_{head}$ and $E_{tail}: \mathbb{R}^{n\times h_{gc}} \rightarrow \mathbb{R}^{n\times h_{bi}}$, we first project $\mathbf{H}_{gc}$ into head ($\mathbf{H}_{head}$) and tail ($\mathbf{H}_{tail}$) features:
\[ \mathbf{H}_{\{head,tail\}} = E_{\{head,tail\}}(\mathbf{H}_{gc}) \in \mathbb{R}^{n\times h_{bi}} \]
Then, for $\forall 1\leq i,j\leq n$, the biaffine attention is performed between the head features of the i\textsuperscript{th} token $\mathbf{h}_{head}^{(i)}$ and the tail features of the j\textsuperscript{th} token $\mathbf{h}_{tail}^{(j)}$, producing $\mathbf{s}_{i,j}\in \mathbb{R}^2$ representing the probability that an edge exists between the i\textsuperscript{th} and j\textsuperscript{th} token:
\[ \mathbf{s}_{i,j} = \text{softmax}(\text{Biaff}(\mathbf{h}_{head}^{(i)}, \mathbf{h}_{tail}^{(j)})) \]
\[ \text{Biaff}(\mathbf{x},\mathbf{y}) := \mathbf{x}^T\mathbf{U}_1\mathbf{y} + \mathbf{U}_2(\mathbf{x}\oplus \mathbf{y}) + \mathbf{b} \]
As $\mathbf{U}_1\in\mathbb{R}^{2\times h_{bi}\times h_{bi}}$, $\mathbf{U}_2\in\mathbb{R}^{2\times 2h_{bi}}$ and $\mathbf{b}\in\mathbb{R}^2$ are trainable parameters, $\oplus$ denotes concatenation. Combining all scores by $\mathbf{Y}_{gc} = (\mathbf{s}_{i,j})_{(1\leq i,j\leq n)}\in \mathbb{R}^{n\times n\times 2}$, we now have a table describing the edge existence likelihood between any two tokens.

At training, we first obtain the boolean gold table $\mathbf{G}_{gc}\in \mathbb{B}^{n\times n}$, for every connected node pair in the query graph, the element in $\mathbf{G}_{gc}$ corresponding to any pair of tokens belonging to the mentions of the two nodes respectively is set to 1 (resulting in several rectangles of 1s). Also, we prefix the question with a special [CLS] token and connect it with the target node to represent the "select" edge; for ASK queries without target nodes, a [SEP] token is suffixed and connected with [CLS]. Note that since the graph structure is undirected, $\mathbf{G}_{gc}$ is a symmetric matrix. An example of $\mathbf{G}_{gc}$ can be found in Figure \ref{fig:graph_structure_generation}. With $\mathbf{G}_{gc}$, we can train the table-filler by $\ell_{tb}$:
\[ \ell_{tb} = -\frac{1}{n^2} \sum\limits_{i=1}^n\sum\limits_{j=1}^n log(\mathbf{Y}_{gc}[i;j;\mathbf{G}_{gc}[i;j]]) \]
Following \citealp{wang2021unire}, we also introduce $\ell_{sym}$ to grasp the table symmetry. Finally, we optimize $\ell_{gc} = \ell_{tb} + \ell_{sym}$ to train a model for GC.
\[ \ell_{sym} = \frac{1}{n^2} \sum\limits_{i=1}^n\sum\limits_{j=1}^n\sum\limits_{k=1}^2 |\mathbf{Y}_{gc}[i;j;k]-\mathbf{Y}_{gc}[j;i;k]| \]

At inference, for each pair of nodes given by NE, we average the rectangle area in $\mathbf{Y}_{gc}$ corresponding to the mentions of the node pair as its edge existence probability. The node pairs with a probability higher than 0.5 are connected. This threshold is selected intuitively to denote an edge is more likely to exist against to not exist, though we argue that the prediction is insensitive to any threshold in reasonable range (e.g. 0.3$\sim$0.7).

\begin{figure}[t]
    \centering
    \includegraphics[width=.92\linewidth]{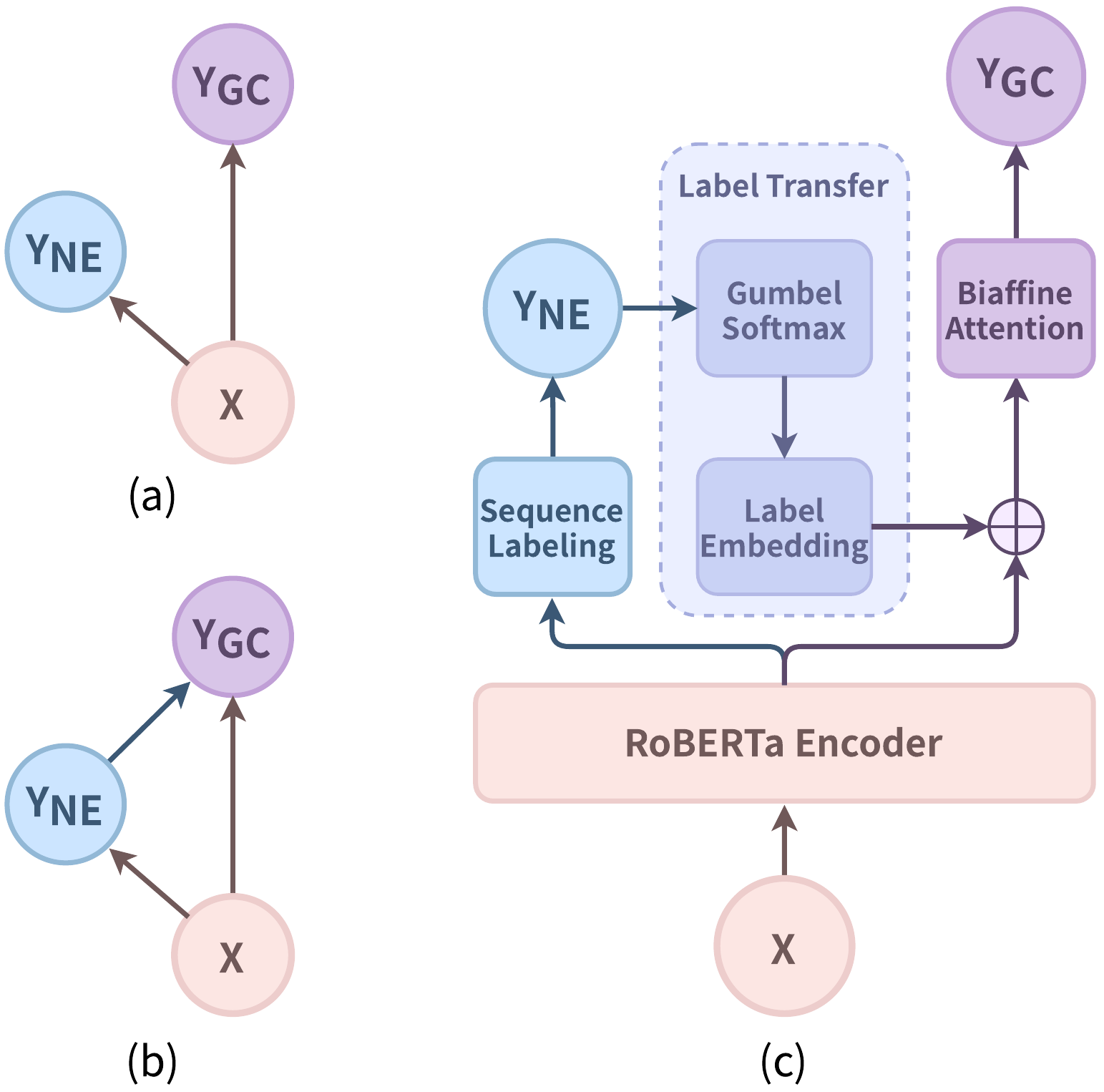}
    \caption{Modelling NE and GC with (a) and without (b) causality, as X, Y\textsubscript{NE}, and Y\textsubscript{GC} denotes question, NE predictions, and GC predictions. Model (c) learns the causal effects by a label transfer module.}
    \label{fig:causal_graphs}
\end{figure}

\subsection{Causal Modelling NE and GC}
\label{sec:causal_modelling}
Up to now, NE and GC are treated as separate tasks that fail to model the intrinsic causal effects between them (e.g. edges in Y\textsubscript{GC} only exist between the mentions detected in NE). Here, we model such causality by a mediation assumption in Figure \ref{fig:causal_graphs}(b) denoting the causal dependence of GC on both question and NE prediction by edge X\textrightarrow Y\textsubscript{GC} and Y\textsubscript{NE}\textrightarrow Y\textsubscript{GC} respectively. To grasp this causal graph, we devise a label transfer \citep{chen-etal-2020-exploring-logically} module to enable the transfer of NE predictions to GC, i.e. representing Y\textsubscript{NE}\textrightarrow Y\textsubscript{GC}, in Figure  \ref{fig:causal_graphs}(c).

In detail, we sample NE predictions $\widetilde{\mathbf{Y}_{ne}}$ by gumbel softmax \cite{nie2018relgan} with $\boldsymbol{g}\!\!\sim$Gumbel(0,1) and temperature $\tau$.
\[ \widetilde{\mathbf{Y}_{ne}} = \text{softmax}((\mathbf{Y}_{ne}+\boldsymbol{g})/\tau) \in \mathbb{R}^{n\times |L|} \]
$\widetilde{\mathbf{Y}_{ne}}$ is then embedded by label embedding $\mathbf{W}_{le}\in \mathbb{R}^{|L|\times h_{le}}$ and concatenated with $\mathbf{H}_{rb}$ to form $\mathbf{H}_{gc}$ in Section \ref{sec:gc} with $h_{gc}\!=\!h_{rb}+h_{le}$:
\[ \mathbf{H}_{gc} = \mathbf{H}_{rb} \oplus (\widetilde{\mathbf{Y}_{ne}} \mathbf{W}_{le})\in \mathbb{R}^{n\times h_{gc}} \]
Now, by minimizing $\ell_{gsg}\!=\!\ell_{ne}\!+\!\ell_{gc}$, a joint model for NE and GC can be obtained. In this model, GC receives NE labels to learn the causal effects from NE, while NE gets feedback through differentiable label transfer to further aid GC decision. In this sense, our model improves the integrity of graph structure generation compared with separately modelling each subtask or simple multitasking.

\begin{figure}[t]
    \centering
    \includegraphics[width=.999\linewidth]{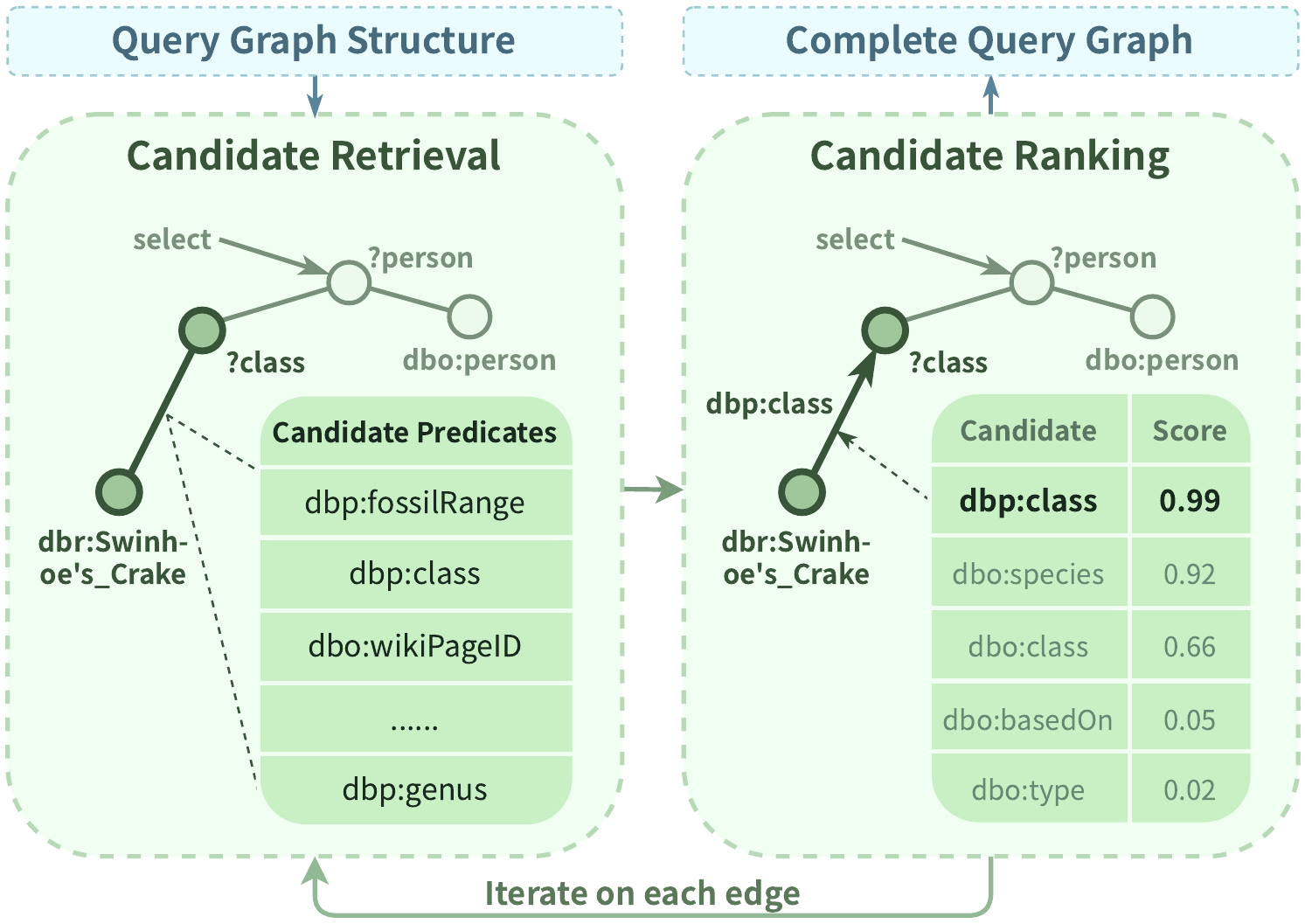}
    \caption{Candidate retrieval and ranking framework for relation extraction.}
    \label{fig:re_overview}
\end{figure}

\section{Relation Extraction (RE)}
\label{sec:re}
As shown in Figure \ref{fig:re_overview}, relation extraction (RE) conducts candidate retrieval and ranking in turn for each edge in graph structure $S$ to decide its predicate. For a question $q$, an edge $e$ connecting nodes $n_1$ and $n_2$ with mention $m_1,\!m_2$ respectively, candidate retrieval recalls a set of predicates $P$ that can be bound to $e$. Note that unlike $e$, each predicate in $P$ is directional. Then, candidate ranking \texttt{Rank}($P$,$q$,$m_1$,$m_2$) gives each predicate a score. This section details this procedure.
\paragraph{Candidate Ranking}
\label{sec:re_cand_rank}
For each $p_i\in P$, we encode it together with $q,\!m_1,\!m_2$ by a RoBERTa encoder and pool them to $0\!\leq\! s_i\!\leq\! 1$ to score the predicate. If the direction of $p_i$ is $n_1\!\!\rightarrow\! n_2$, we join $q,m_1,m_2,p_i$ sequentially by [SEP] token as model input; otherwise (direction $n_2\!\!\rightarrow\! n_1$), the join order is $q,m_2,m_1,p_i$. By giving $s_i$ to each candidate, we can get the most proper predicates for $e$ by selecting those with highest scores. More details on training the ranking model can be found in Appendix \ref{sec:appendix_re_ranking}.
\paragraph{Candidate Retrieval}
\citealp{zhang2021namer} proposed a straightforward way to retrieve candidates: if either $n_1$ or $n_2$ is a non-variable node, the predicates around that node in KB are viewed as candidates; otherwise, they trace $n_1$ or $n_2$ in other graph edges with non-variable nodes and view the predicates k-hop away from that node in KB as candidates (e.g. predicates 2-hop away from \textit{dbo:person} are candidates for $?class\text{-}?person$ in Figure \ref{fig:re_overview}). We view this as the baseline in latter experiments.

\begin{algorithm}[t]
\caption{BeamSearchRE}\label{alg:re_bs}
\LinesNumbered
\small

\SetKwFunction{FR}{Sample}
\SetKwFunction{FI}{inference}
\SetKwFunction{FG}{getNodePairsFromGraph}
\SetKwFunction{FC}{Retrieve}
\SetKwFunction{FCR}{Rank}

\KwIn{Question $q$, Query graph structure $S$, beam width $b$}
\KwOut{A beam of query graphs $B$}
$B\leftarrow\{\{\}\}$;\tcp*[h]{Start with an empth graph}

$S_{pend}\leftarrow S$;\tcp*[h]{All edges are pending}

\While{$S_{pend}\ne\varnothing$}{
    $B'\leftarrow\{\}$\;
    \tcp{Select a pending edge}
    $e=(n_1,n_2)\leftarrow$\FR$(S_{pend})$\;
    \For{$G\in B$
    }{
    $P\leftarrow$\FC$(G, n_1, n_2)$\;
    \tcp{$n_1/n_2$ has mention $m_1/m_2$}
    $C=\{(p_i,s_i)\}\leftarrow$\FCR$(P, q, m_1, m_2)$\;
    \tcp{Extend previous beams}
    \For{$(p_i,s_i)\in C$}{
        $B'\leftarrow B'\cup \{G \cup \{(n_1, n_2, p_i, s_i)\} \}$;
    }
    }
    $B\leftarrow B'.\text{topk}(b)$;\tcp*[h]{Set up new beams}
    
    \tcp{Mark \textit{e} as determined}
    $S_{pend}\leftarrow S_{pend}\setminus \{e\}$\;
}
\end{algorithm}

However, this results in a candidate scale O($n^k$)\footnote{n is the node degree in KB, k is the edge number in $S$}, making it unscalable to multi-hop queries (k$\uparrow$) and large KBs (n$\uparrow$). Here, we propose Algorithm \ref{alg:re_bs} to limit the scale to O(n). We start by selecting an edge between $n_1^a$ and $n_1^b$ containing a non-variable node (e.g. edge \textit{?class}-\textit{dbr:Swinhoe's\_Crake} in Figure \ref{fig:re_overview}), retrieving all adjacent predicates of that node in KB and use \texttt{Rank} to select the most proper predicate $p_1$ (e.g. dbp:named\_by) of score $s_1$, this forms a subgraph $G$=\{($n_1^a$,$n_1^b$,$p_1$)\} with only one edge whose score is $s_1$. Then, we sample another edge between $n_2^a$ and $n_2^b$ (e.g. \textit{?class}-\textit{?person}) and retrieve its candidates $P$ based on $G$ (e.g. $G$ already entails \textit{?class}=\textit{dbr:bird}, so all neighbors of \textit{dbr:bird} forms $P$), this process is denoted as \texttt{Retrieve}($G,n_2^a,n_2^b$). Now, we use \texttt{Rank} to select $p_2$ of score $s_2$ from $P$, add ($n_2^a$,$n_2^b$,$p_2$) to subgraph $G$ and update its score as $s_1*s_2$. Repeating this loop until all edges are bound with a predicate, we finally form a query graph.

Note that for each edge, the candidate scale given by \texttt{Retrieve} is O(n), since it is always among the neighbors of one or several KB nodes. Also, to improve the recall of query graphs, this process can trivially be extended as a beam search with each step maintaining a beam of subgraphs $B$, ordering each subgraph by $\prod_i s_i$ as in Algorithm \ref{alg:re_bs}.

\begin{table}[t]
\centering
\resizebox{.99\columnwidth}{!}{
\begin{tabular}{c l c c c}
    \toprule
    \bfseries Type & \bfseries Methods & \bfseries P & \bfseries R & \bfseries F1\\
    \cmidrule(lr){1-5}
    \multirow{2}{*}{\bfseries I} &
    NSQA \citep{kapanipathi2021leveraging} & .448 & .458 & .445 \\
    & EDGQA \citep{hu2021edg} & .505 & .560 & .531 \\
    \cmidrule(lr){1-5}
    \multirow{4}{*}{\bfseries II} &
    QAmp \citep{vakulenko2019message} & .250 & .500 & .330 \\
    & NAMER \citep{zhang2021namer} & .438 & .438 & .435 \\
    & STaG-QA \citep{ravishankar2021two} & \bfseries.745 & .548 & .536 \\
    & \bfseries Crake (ours) & .722 & \bfseries.731 & \bfseries.715 \\
    \bottomrule
\end{tabular}}
\caption{End-to-end performance on LC-QuAD 1.0 test set. I/II stands for methods with/without aux tools. We re-implement NAMER since its results on LC-QuAD is not provided; however, NAMER suffers from severe timeout issues on DBpedia to limit its performance, so we restrict each candidate query to run at most 45s in practice (which already requires $\sim\!$15h for a complete evaluation run).}
\label{tab:e2e_eval}
\end{table}

\section{Experiments}
\label{sec:exp}
\paragraph{Dataset} We adopt LC-QuAD 1.0 \citep{trivedi2017lc}, a predominant open-domain English KBQA benchmark based on DBpedia \citep{auer2007dbpedia} 2016-04, to test the performance of our system. We randomly sample 200 questions from train data as dev set and follow the raw test set, resulting in a 4800/200/1000 train/dev/test split. More details on the dataset can be found in Appendix \ref{sec:appendix_dataset}. Like \citealp{zhang2021namer}, we do not experiment on multiple datasets due to the high annotation cost involved, however, we conduct no dataset-specific optimizations in this work, so we consider the large improvements on LC-QuAD and detailed discussions sufficient to prove our effectiveness.
\paragraph{Annotation}
\label{sec:annotation}
We annotate the dataset with the mention of each node in query graph, e.g. the mention "class" and "person" for the node \textit{?class} and \textit{dbo:person} respectively in Figure \ref{fig:two_stage}. With the annotation, we obtain the gold data ($G_{ne},\!G_{gc}$) to train our models. Appendix \ref{sec:appendix_annotation} details the annotation process. 

\paragraph{Baselines}
We evaluate our method against existing works both with and without auxiliary tools. With aux tools, \citealp{kapanipathi2021leveraging} constructs query graphs based on the AMR of questions; \citealp{hu2021edg} designs rules on constituency tree to aid query graph formation. For independent pipelines without aux tools, \citealp{vakulenko2019message} parses URI mentions from the question to match with KB via confidence score passing; \citealp{ravishankar2021two} combines a generative graph-skeleton decoder with entity and relation detector to form a query; \citealp{zhang2021namer} co-trains a pointer generator with the node extractor to build a query graph, it's worth to note that this work also requires the node-to-mention \nameref{sec:annotation} for training.

\paragraph{Setup}
We utilize the RoBERTa-large released by huggingface \citep{wolf-etal-2020-transformers} as our encoder. All experiments are averaged on two runs on an NVIDIA A40 GPU. For the GSG model, we train for at most 500 epochs (\textasciitilde6 GPU-hours) and report the best checkpoint on dev set; for the RE model, we train for 20 epochs (\textasciitilde16 GPU-hours) and report the final checkpoint. For hyperparameters, we find no apparent performance variance on dev set as long as the values are in reasonable range (e.g. $64\!\leq\! h_{le}\!\leq\!1024$, $1e\text{-}6\!\leq\! lr_{gsg}\!\leq\!2e\text{-}5$) so no further tuning is involved. See the full setting in Appendix \ref{sec:appendix_hyper}.

\begin{table*}[t]
\centering
\resizebox{1.85\columnwidth}{!}{
\begin{tabular}{l l c c c c c c c c}
    \toprule
    \multirow{2}{*}{\bfseries Methods} &
    \multirow{2}{*}{\bfseries Decoder Parameters} &
    \multicolumn{3}{c}{\bfseries NE Accuracy} &
    \multicolumn{2}{c}{\bfseries GSG Accuracy} &
    \multicolumn{3}{c}{\bfseries End-to-end}\\
    \cmidrule(lr){3-5}
    \cmidrule(lr){6-7}
    \cmidrule(lr){8-10} & &
    P & R & F1 &
    EM & Actual &
    P & R & F1\\
    \cmidrule(lr){1-10}
    Seq2seq & 76.67M\space($\times1$) &  .895 & .901 & .897 & .695 & .768 & .653 & .674 & .654\\
    TF  & 0.66M\space\space\space($\times1/100$) & .895 & .901 & .897 & .728 & .795 & .655 & .674 & .657\\
    TF+SMTL & 0.66M\space\space\space($\times1/100$) & .901 & .904 & .902 & .735 & .805 & .665 & .684 & .667\\
    \cmidrule(lr){1-10}
    TF+Causal & 3.03M\space\space\space($\times1/25$) & \bfseries.909 & \bfseries.914 & \bfseries.911 & \bfseries.755 & \bfseries.828 & \bfseries.677 & \bfseries.696 & \bfseries.680\\
    \bottomrule
\end{tabular}}
\caption{Experiments on table-filling and causal-modelling. Seq2seq and TF adopt a generative decoder and table-filler in GC respectively, while both deal with NE and GC by separate models. TF+SMTL (simple multitask learning) co-trains NE and GC by directly adding losses without modelling their intrinsic causal effects. TF+Causal denotes our full approach which models the causal effects between NE and GC by label transfer. We report the node-level P/R/F1 in NE, the exact-match (EM) and actual accuracy (that ignores variable mentions in judging accuracy) in GSG, and the overall answer-level P/R/F1 on LC-QuAD 1.0 dev set for comparison.}
\label{tab:ablation}
\end{table*}

\begin{figure}[t]
    \centering
    \includegraphics[width=.999\linewidth]{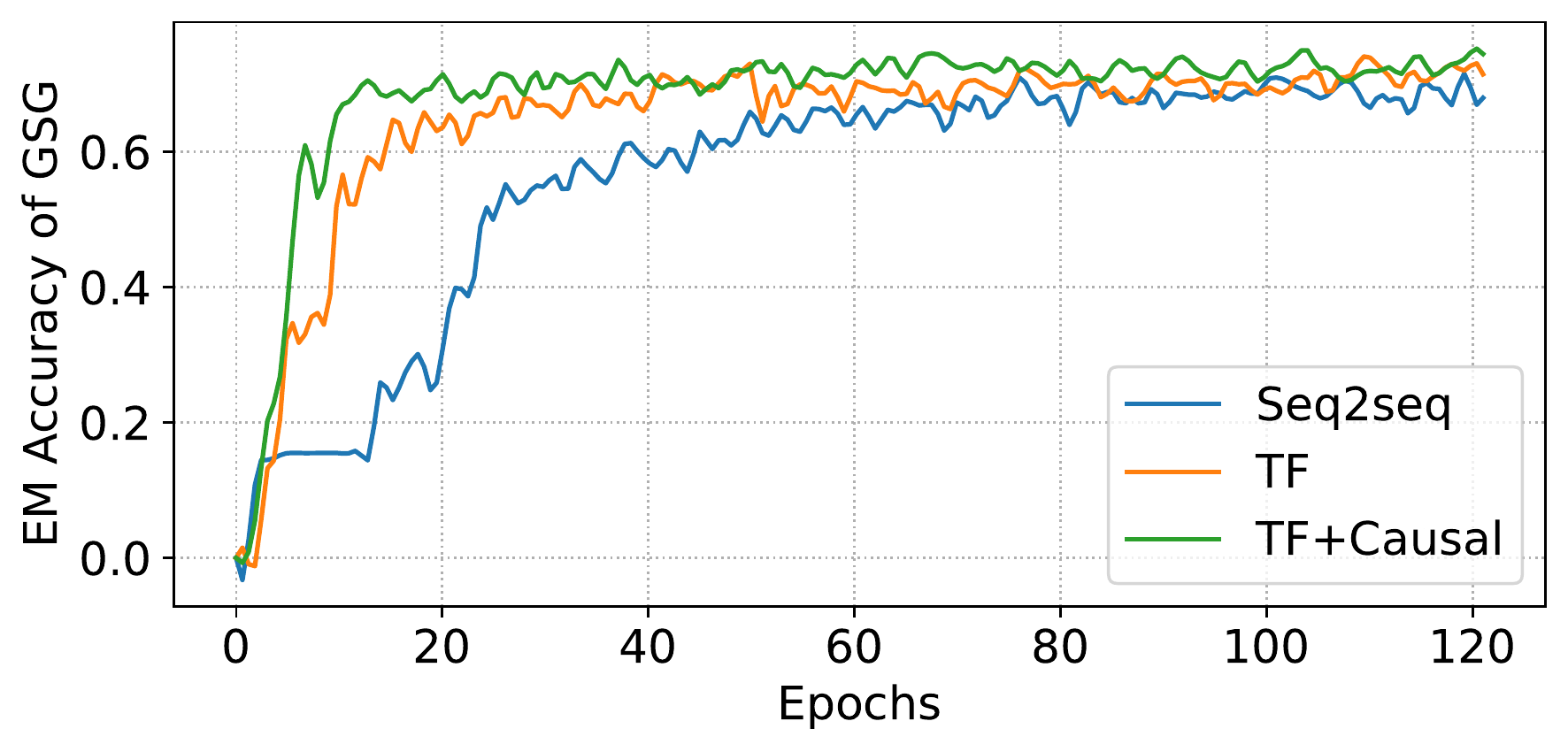}
    \caption{EM accuracy of GSG during training. See the meaning of each series in Table \ref{tab:ablation}.}
    \label{fig:learn_curve}
\end{figure}

\subsection{End-to-end Evaluation}
As shown in Table \ref{tab:e2e_eval}, our method, Crake, outperforms all former methods by a large $\!\sim\!$17\% margin on F1, becoming the new SoTA of LC-QuAD 1.0. Surpassing methods requiring aux tools (I) on all metrics, we present the effectiveness of independent pipelines (II) that avoid cascading errors. Also, we achieve consistent answer precision and recall to surpass other methods in II on F1, showing the superiority of our pipeline design, which is further discussed in the sections below.

\subsection{Effects of Tabel-Filling}

As explained in Section \ref{sec:gc}, modelling GC as a sequence-generation causes a few issues that can be overcome by table-filling. Specifically, the sequence ambiguity confuses the learning process and requires large decoders to grasp the sequence generation policy, which may slow down the convergence. Besides, the exposure bias harms the decoding accuracy of the model at inference. This section, we try to verify such effects by experiments. To enable the comparison with sequence-generation, we construct a generative decoder as in \citealp{zhang2021namer} as the baseline, which sequentially generates the connected node pairs in the graph structure to represent the edges. We train the generative model under the same settings (e.g. learning rate, warmup, epochs, etc.), resulting in the performance of \texttt{Seq2seq} in Table \ref{tab:ablation}.

\begin{figure}[t]
    \centering
    \includegraphics[width=.93\linewidth]{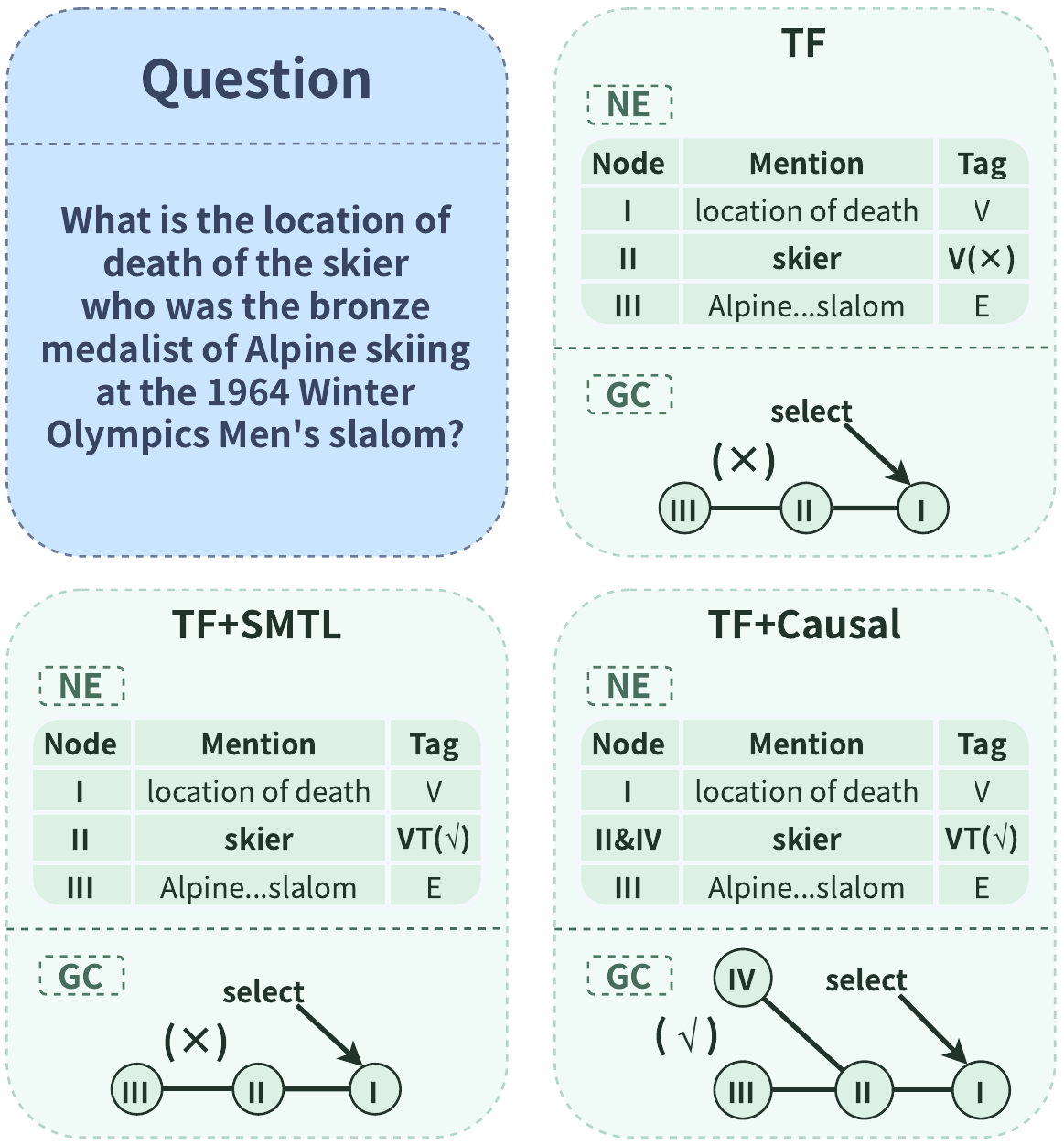}
    \caption{Case study on the effects of causal-modelling.}
    \label{fig:case_study}
\end{figure}

Comparing with the table-filling model (i.e. \texttt{TF} in Table \ref{tab:ablation}), \texttt{Seq2seq} comes short in the accuracy of graph structure, indicating the negative effects of the exposure bias on predicting accuracy. Meanwhile, \texttt{TF} requires only 1/100 of \texttt{Seq2seq}'s parameters to achieve comparable or better results, we attribute this to the removal of sequence ambiguity which frees the model from acquiring the complex and ambiguous scheme of sequence-generation. This speculation is further verified in Figure \ref{fig:learn_curve}, in which \texttt{TF} converges distinctly quicker than \texttt{Seq2seq} since the simultaneous decision of all edges is well-defined and easier to learn. Thus, compared with sequence-modelling, handling GC via table-filling reduces model size and boosts training, which is essential for real-world applications.

\subsection{Effects of Causal-Modelling}

We propose a joint model to learn the NE-GC causalities in Section \ref{sec:causal_modelling}, to discuss its effects, we compare it with two alternatives in Table \ref{tab:ablation}: 1) using two separate models for NE and GC (\texttt{TF} in Table \ref{tab:ablation}) like \citealp{ravishankar2021two}, 2) co-training NE and GC by sharing encoder and adding losses (\texttt{TF+SMTL} in Table \ref{tab:ablation}) like \citealp{shen2019multi}. As shown, co-training consistently surpasses separate models by grasping the shared knowledge between NE and GC, nevertheless, our causal-modelling approach (\texttt{TF+Causal}) further outperforms co-training. In detail, though \texttt{TF+Causal} has similar results with \texttt{TF+SMTL} in NE, it achieves better accuracy for overall GSG (NE+GC) and excels in end-to-end metrics. Therefore, we infer that causal-modelling improves the integrity of the GSG stage by expressing the internal causalities between its subtasks. To better understand this, we perform a case study in Figure \ref{fig:case_study}, in which \texttt{TF} fails to realize that "skier" also corresponds to a type node; in contrast, \texttt{TF+SMTL} extract all nodes correctly by learning both NE and GC labels, but it still fails in generating a correct graph structure. Finally, \texttt{TF+Causal} utilizes the VT tag of "skier" in NE predictions and correctly connects the II-IV edge in GC. Thus, Figure \ref{fig:case_study} demonstrates the usage of causal effects to reach higher accuracy in GSG.

\subsection{Analysis on Beam-Search RE}

\begin{table}[t]
\centering
\resizebox{1.\columnwidth}{!}{
\begin{tabular}{l c c c c c c}
    \toprule
    \multirow{2}{*}{\bfseries Methods} & 
    \multicolumn{3}{c}{\bfseries Accuracy} &
    \multicolumn{3}{c}{\bfseries Efficiency}\\
    \cmidrule(lr){2-4}
    \cmidrule(lr){5-7} & \bfseries P & \bfseries R & \bfseries F1 & \bfseries 1-hop & \bfseries 2-hop & \bfseries 3-hop\\
    \cmidrule(lr){1-7}
    Baseline & .560 & .566 & .556 & \bfseries0.12s & 42.4s & 84.2s\\
    BeamSearch & \bfseries.677 & \bfseries.696 & \bfseries.680 & \bfseries0.12s & \bfseries1.06s & \bfseries2.72s\\
    \bottomrule
\end{tabular}}
\caption{Performance comparison between our beam-search RE algorithm and its baseline in Section \ref{sec:re}. Accuracy refers to the answer-level P/R/F1, efficiency is measured by the average run time on 1/2/3-hop queries.}
\label{tab:re_ablation}
\end{table}

In this section, we compare our beam-search RE algorithm with its baseline. As stated in Section \ref{sec:re}, by alternately performing retrieval and ranking on each edge (rather than retrieving the candidates of every edge before ranking), our approach lowers the scale of candidate predicates on multi-hop queries to get better efficiency, which is verified in Table \ref{tab:re_ablation}. In detail, \texttt{BeamSearch} costs substantially less time than \texttt{Baseline} in 2 and 3-hop queries (note that for 1-hop queries, two methods reduce to a same process with similar time costs). Since \texttt{BeamSearch} only operates on the neighbors of certain KB nodes, it avoids the retrieval of 2-hop neighbors, which requires considerable time on DBpedia, to improve efficiency. In addition, by pruning off useless candidates in \texttt{Baseline}, \texttt{BeamSearch} also achieves higher overall KBQA accuracy in Table \ref{tab:re_ablation}. 
Therefore, Algorithm \ref{alg:re_bs} transcends previous methods to reveal an efficient and accurate solution for ranking-based RE scalable to KB size and query complexity.

\section{Related Works}
\paragraph{KBQA via Semantic Parsing} A mainstream to solve KBQA is semantic parsing \citep{yih2016value} which converts a question to a KB query to get answers. Due to the graph-like structure of KB queries, prior works construct query graphs to represent queries in semantic parsing. Among them, some works \citep{zafar2018formal, chen2021formal} only focus on predicting the graph structure given node inputs. To perform end-to-end QA, \citealp{hu2017answering} leverages the dependency parsing tree to match KB subgraphs for answers; \citealp{kapanipathi2021leveraging} builds the query graph by transforming and linking the AMR \citep{banarescu2012abstract} of the question; \citealp{hu2021edg} uses the constituency tree to compose an entity description graph representing the query graph structure. Requiring aux tools or data structures, these works may be subjected to cascading errors. \citealp{yih2015semantic} overcomes this by an independent stage-transition framework to generate the query graph, \citealp{hu2018state} extends the transitions to express more complex graphs. Besides, \citealp{zhang2021namer} adopts a pointer generator to decode graph structure, \citealp{ravishankar2021two} generates the query skeleton by a seq2seq decoder. Unlike these methods that model the query graph as a sequence (by state-transition or generative decoder), we decode all edges at once via a table-filler in graph structure generation.

\paragraph{Modelling causal effects} Causality occurs in various deep-learning scenarios between multiple channels or subtasks, existing works models the causality for better performance. \citealp{niu2021counterfactual} mitigates the false causal effects in VQA \citep{antol2015vqa} to overcome language bias; \citealp{zeng2020counterfactual} dispels the incorrect causalities from different input channels of NER by generating counterfacts. \citealp{chen-etal-2020-exploring-logically} utilizes the inter-subtask causalities to improve multitask learning for JERE \citep{li2014incremental}, ABSA \citep{kirange2014aspect}, and LJP. Unlike them, we formulate and utilize the internal causal effects in KBQA.


\section{Conclusion}
In this work, we formalize the generation of query graphs in KBQA by two stages, namely graph structure generation (GSG) and relation extraction (RE). In GSG, we propose a table-filling model for graph composition to avoid the ambiguity and bias of sequence-modelling, meanwhile, we encode the inherent causal effects among GSG by a label-transfer block to improve the stage integrity. In RE, we introduce an effective beam-search algorithm to retrieve and rank predicates in order for each edge, which turns out to be scalable for large KBs and multi-hop queries. Consequently, our approach substantially surpasses previous state-of-the-arts in KBQA, revealing the effectiveness of our pipeline design. Detailed experiments also validate the effects of all our contributions.


\section{Limitation}
Admittedly, our approach endures certain limitations as discussed below.
\paragraph{Query Expressiveness} Like most semantic parsing systems, we fail to cover all the operations of SPARQL, limiting our capability to compose queries with complex \texttt{filter} or property path. For the conciseness of our system, we only focus on constructing triples in the multi-hop query graph in this paper, while we plan to incorporate more functions into Crake in the future to improve the expressiveness of the system.
\paragraph{Annotation Cost} Training models with node mentions require expensive manual annotations, which is impractical for us to conduct on every popular KBQA dataset. As explained in Section \ref{sec:exp}, without data-oriented optimization, we believe the significant gain presented adequate to verify our contributions. Further, we expect to extenuate such costs in two directions for the future: 1) some modules of our framework (e.g. NE) is generalizable to other English questions, gifting it the potential to be transferred to other datasets without re-training; 2) few-shot \citep{wang2020generalizing} and active \citep{aggarwal2014active} learning techniques aids the model to reach competitive performance with a small portion of annotated data, which can be explored in our framework to reduce annotation cost.

\section*{Acknowledgements}

This work was supported by National Key R\&D Program of China (2020AAA0105200) and NSFC under grant U20A20174. The corresponding author of this work is Lei Zou (zoulei@pku.edu.cn). We would like to thank Zhen Niu and Sen Hu for their kind assistance on this work. We also appreciate anonymous reviewers for their valuable comments and advises.

\bibliography{anthology,custom}
\bibliographystyle{acl_natbib}

\appendix

\section{Details in Entity and Type Linking}
\label{sec:appendix_linking}
We link each non-variable node to a KB entry by its mention. For entity nodes, we directly link it to an entity with the same name as its mention if such entity exists in the KB (e.g. link mention "New York" to dbr:New\_York); otherwise, we recall entities by DBpedia Lookup\footnote{http://wiki.dbpedia.org/projects/dbpedia-lookup} and further prioritize ones whose lower-cased name is the same as the lower-cased mention (e.g.dbr:new\_york). Then, the prioritized entity with the highest lookup score is linked to the node; if no entity is prioritized, the entity with the highest score is selected.

For type nodes, we build a dictionary $D$ based on the mention-type pairs (e.g. authors-dbo:Writer) in train data and directly use the link result from $D$ if the mention exists in $D$. Otherwise, we singularize and capitalize the mention to construct an URI with prefix \textit{dbo} (e.g. bands$\rightarrow$dbo:Band), if this URI presents in the KB, the type node is linked to this URI. If no entry is found for either an entity or a type afterall, we simply discard the node from our query graph.

Note that although we involve no extra disambiguation step, DBpedia Lookup itself has certain mention-level disambiguation abilities to refine mention-relevant candidates. Admittedly, sentence context also contributes to a precise linking decision, leaving such context-level disambiguation a future direction to improve our work.

\section{Training Details of the Candidate Ranking Model}
\label{sec:appendix_re_ranking}
We mainly follow NAMER \citep{zhang2021namer} in training the candidate ranking model mentioned in Section \ref{sec:re_cand_rank}. Basically, the positive and negative training samples are obtained from the gold query. For instance, in Figure \ref{fig:two_stage}, we obtain the candidates between ?class ($m_1$="class") and ?person ($m_2$="person") by constructing \textit{"select ?r \{ ?person dbp:type dbo:person. ?class ?r ?person. dbr:Swinhoe's\_Crake dbp:class ?class \}"} and \textit{"select ?r \{ ?person dbp:type dbo:person. ?person ?r ?class. dbr:Swinhoe's\_Crake dbp:class ?class \}"} with query results $P_{p}$ and $P_{r}$ respectively. Let $p^*\!\!=\!\!dbp\!:\!named\_by$ be the correct predicate, we collect model inputs $\{ (q,m_1,m_2,p^*) \}$ as a positive sample (i.e. of label 1) and $\{ (q,m_1,m_2,p_i) | p_i\in P_{pos}\backslash\{p^*\} \cup \{ (q,m_2,m_1,p_i) | p_i\in P_{rev}\}$ as negative samples (i.e. of gold label 0).

Further, we follow the augmentation process in NAMER to learn the effects of mention order on model predictions. Specifically, we add $\{ (q,m_1,m_2,p_i) | p_i\in P_{rev}\backslash\{p^*\} \cup \{ (q,m_2,m_1,p_i) | p_i\in P_{pos}\}$ to negative samples when training. With the aforesaid process repeated on each query graph edge, we get the full training samples to train a ranking model.

Besides, similar with NAMER, we observe a performance decay when forcibly co-training RE and GSG module, in this regard, we leave RE a separate module alongside GSG in the system. As discussed in NAMER, the different input channels between RE and GSG may result in unequal semantic spaces for the model. Thus, despite the causal association between RE and GSG, we conjecture that the model fails to acquire beneficial causalities between incompatible semantic spaces.

\section{Details of the Dataset}
\label{sec:appendix_dataset}
LC-QuAD 1.0 is an English open-domain KBQA dataset widely used to evaluate KBQA systems. With a GPL-3.0 licence, this dataset is intended for training and testing models to answer a question via querying the knowledge base, permitting modifications on the dataset for experiments, which is consistent with the way we use the dataset (annotating node mentions for each data entry, train several models for KBQA on the train data and test the system performance on the test data).

Due to the nature of KBQA tasks, LC-QuAD 1.0 involves questions about certain real-world entities usually including persons, organizations or objects (which is exactly the conditions where KBQA is applied to real-world applications). However, most information about the individuals (e.g. name, team, etc.) are publicly available (since the dataset utilizes DBpedia as background KB while DBpedia mainly collects data from publicly available Wikipedia). Further, when annotating the dataset, we perform a brief manual check on potential offensive or biased contents, to the best of our efforts, we find no apparent offensive hints in the questions and SPARQL queries. Hence, we believe that LC-QuAD 1.0 under intended KBQA use has minor potential to offend others or cause privacy issues.

\section{Details of Data Annotation}
\label{sec:appendix_annotation}
\paragraph{Annotation Guidelines}
We adopt the same annotation format as \citealp{zhang2021namer} to annotate the LC-QuAD 1.0 dataset. Specifically, for each node in the query graph corresponding to the SPARQLs in the dataset, the mention of such node in the question is annotated. All annotated mentions are required as whole-words (e.g. including the 's' for plural words), the mention is left as "None" when no mention of a node can be found. There are certain cases where multiple mentions co-refer a node, we encourage annotators to choose a mention containing more concrete semantics, while all of these mentions are acceptable (e.g. for the question "Who is Jack's dad?", both "Who" and "dad" are correct mentions but the latter is encouraged since it indicates more semantics of the node). We provide a detailed guideline\footnote{See the guideline in supplementary materials.} to annotators with extra discussions on marginal cases to further aid the annotation. Also, we discuss the potential risks and the overall usage of such annotations to get agreements from the annotators in the guideline.
\paragraph{Annotation Process} We recruit 9 annotators with necessary background knowledge from school, consisting of 5 undergraduate and 4 graduate students, to fulfil the annotation task. By completing the annotation, we provide essential payments for each annotator. Finally, we use a script to auto-check the collected annotations and perform basic corrections (e.g. align all mentions to whole-words).

\section{Hyperparameter Settings}
\label{sec:appendix_hyper}
Table \ref{tab:hyper} details our hyperparameter settings.

\begin{table}[h]
\centering
\resizebox{1.\columnwidth}{!}{
\begin{tabular}{c l c}
    \toprule
    \bfseries Name & \bfseries Description & \bfseries Setting\\
    \cmidrule(lr){1-3}
    $h_{rb}$ & Hidden size of the RoBERTa encoder & 1024\\
    $h_{bi}$ & Hidden size of the biaffine model & 256\\
    $h_{le}$ & Dimension of the label embedding & 256\\
    $\tau$ & Gumbel-softmax temperature & 0.05\\
    $optim$ & Optimizer to train both GSG and RE models & AdamW\\
    $\beta_1 / \beta_2$ & Betas of the AdamW optimizer & 0.9 / 0.9 \\
    $wd$ & Weight decay rate of the AdamW optimizer & 1e-5\\
    $lr_{rb}$ & Learning rate of the RoBERTa encoder in GSG & 1e-5 \\
    $lr_{gsg}$ & Learning rate of other parameters in GSG & 5e-5\\
    $batch_{gsg}$ & Batch size of the GSG model & 64\\
    $lr_{re}$ & Learning rate of the RE model & 1e-5\\
    $batch_{re}$ & Batch size of the RE model & 100\\
    $b$ & Beam width in RE & 4\\
    \bottomrule
\end{tabular}}
\caption{Detailed hyperparameter settings in this work.}\label{tab:hyper}
\end{table}

\section{Ethical Statements}
Considering the nature of NLP-based QA systems, our method keeps the risk to output false (e.g. incorrect answers to factoid questions) or biased (e.g. imprecise count of answer numbers) answers, which might cause issues in trustworthy or practical uses. However, we'd like to clarify that this work is intended for discovering more accurate and efficient systems on KBQA regardless of the exact content in a KB, the answers to specific questions given by our method does not reflect the authors' point of view.

\end{document}